\documentclass{article}
\usepackage[dblblindworkshop, final]{neurips_2026}
\workshoptitle{Interpretability as a Science}
\usepackage[utf8]{inputenc}
\usepackage[T1]{fontenc}
\usepackage[expansion=false]{microtype}
\usepackage{graphicx}
\usepackage{booktabs}
\usepackage{amsmath}
\usepackage{amssymb}
\usepackage{amsthm}
\usepackage{mathtools}
\usepackage{xcolor}
\usepackage{hyperref}
\usepackage{placeins}
\usepackage{float}
\usepackage[capitalize,noabbrev]{cleveref}

\theoremstyle{plain}
\newtheorem{theorem}{Theorem}[section]
\newtheorem{proposition}[theorem]{Proposition}

\newtheorem{corollary}[theorem]{Corollary}
\theoremstyle{definition}
\newtheorem{definition}[theorem]{Definition}

\theoremstyle{remark}
\newtheorem{remark}[theorem]{Remark}
\crefname{definition}{Definition}{Definitions}
\crefname{remark}{Remark}{Remarks}
\crefname{assumption}{Assumption}{Assumptions}
\crefname{corollary}{Corollary}{Corollaries}
\title{Pretty Linear Liars: Unregularized Probes Favor\\ Intervention-Unfaithful Features}
\author{%
  Karl Vilhelmsson Emneby \qquad Julian Yocum \qquad Cameron Allen \\
  Center for Human-Compatible AI \\
  UC Berkeley\\
  \texttt{\{karlcal,juliany,camallen\}@berkeley.edu}
}
\begin{document}
\maketitle
\begin{abstract}

Linear probing is a common technique in mechanistic interpretability despite the fact that the decodability of a given feature does not necessarily imply the network uses that feature. Intervention faithfulness, not merely decodability, is the appropriate criterion for whether a feature serves a defined active role in the model. Meanwhile, probe regularization has been used for feature discovery, but the impact of regularization on feature causality remains unclear. We study probe regularization systematically, starting with unregularized ordinary least squares (OLS) regression and considering increasing amounts of regularization via ridge regression. We show theoretically that OLS probes amplify low-variance directions in activation space that have spurious correlations with the target feature, whereas regularization reduces influence from these directions. Empirically, we find in a toy transformer and Llama-2-7B that intervention faithfulness increases up to a point, and then decreases again, and that the optimal regularization strength leads to features bearing striking similarity (in both direction and variance) to those learned by optimizing for intervention outcomes directly. Together, these results provide mechanistic justification for regularizing regression probes.

\end{abstract}
\section{Introduction}
\label{sec:intro}
Linear probing is a standard interpretability tool in which a linear map is fit from a network's internal activations to a quantity of interest. While decoding accuracy is sometimes taken as evidence (but not proof) the network represents the variable in question,
prior work has shown that decoding accuracy is not enough: a variable can be decodable without the probe's direction being actually used by the network. Therefore, causal interventions, not decodability, are the appropriate test of whether the network uses a feature \citep{geiger2021, elazar2021, huang2025}. Whether the probe's own fitting procedure, and regularization in particular, affects the outcome of that test has received little attention.
Some probing studies regularize explicitly, as when \citet{gurnee2023} fit ridge regression probes with the penalty selected by held-out decodability, or implicitly, as when \citet{heinzerling2024} use partial least squares, whose components are selected by covariance with the target. Neither of these papers explains what breaks without regularization beyond stability and generalization concerns.
The closest precedent is in neuroimaging, where \citet{haufe2014} showed that a decoder can weight a channel heavily though it carries no signal. In that setting, however, the weights cannot be tested by intervention. Other work uses unregularized probes and leaves causal tests of the recovered feature vectors out of scope \citep[for example,][]{shai2024}.  
  
We evaluate the question of probe faithfulness by comparing three ways of extracting a feature vector, i.e. a direction in activation space, for the same target. OLS probes and ridge regression probes fit the direction by decodability, differing only in the ridge penalty $\lambda$. Distributed Alignment Search (DAS) fits the direction by gradient descent on intervention outcomes directly \citep{geiger2024das}, so it serves as a causal reference point, a direction chosen for intervention effect rather than for decodability. Comparing where these three land, on the same activations and the same target, isolates the choice of fitting procedure as the variable.
We also establish a theoretical account of why regularization helps. Sampling noise gives even a direction that the network does not use a small nonzero covariance with the target. OLS is known to amplify noise most strongly along directions where values (i.e. activations) vary least \citep{hoerl1970}. Our theoretical contribution is to show why this can lead to high decodability without intervention faithfulness. Ridge regression mitigates this problem by suppressing the amplification of noise along low-variance directions (\cref{sec:theory}). Beyond this theoretical account, we present empirical results on a toy transformer \citep{shai2024} and Llama-2-7B~\citep{touvron2023}. We find that OLS probes decode the target but are indeed intervention-unfaithful (on the toy transformer, even with exhaustive data). Intervention-selected ridge regression probes are largely faithful and point in high-variance directions. The DAS directions align closely with the ridge regression probe directions. This leads us to suggest that, for decodability-based regression probing, the regularization coefficient be chosen by a sweep that maximizes intervention faithfulness.
\section{Setup}
\label{sec:setup} We first define regression probes and intervention faithfulness, then describe our two testbeds and the interventions we use to test each probe's direction on them.
\subsection{Probes, directions, and intervention faithfulness}
\begin{definition}[Regression probe]
\label{def:probe}
Given activations $a \in \mathbb{R}^{d}$ at a fixed site of a neural network and a continuous target variable $f \in \mathbb{R}^{k}$, a \textbf{regression probe} is a linear map that produces a prediction $\hat{f} = W^\top a + b$, where $W \in \mathbb{R}^{d \times k}$ is the probe weight matrix and $b \in \mathbb{R}^k$ is a bias vector.
Given $n$ activation--target pairs $(A, F)$ with $A \in \mathbb{R}^{n \times d}$ and $F \in \mathbb{R}^{n \times k}$, the \textbf{OLS probe} fits $W$ and $b$ by minimizing
\begin{equation}
\mathcal{L}_{\text{OLS}} = \|AW + \mathbf{1}b^\top - F\|_F^2 ,
\end{equation}
where $\|\cdot\|_F$ is the Frobenius norm, and the \textbf{ridge regression probe} adds an $L_2$ penalty on $W$ (but not $b$):
\begin{equation}
\mathcal{L}_{\text{ridge}} = \mathcal{L}_{\text{OLS}} + \lambda \|W\|_F^2 ,
\end{equation}
for regularization strength $\lambda > 0$.
The optimal bias is always $b = \bar{f} - W^\top \bar{a}$ (the sample means), so fitting with a bias is equivalent to mean-centering $A$ and $F$ before fitting $W$. We treat activations and targets as centered from here on.
We say a target variable is \textbf{linearly decodable} at a site if a probe achieves low mean squared error (MSE), and refer to the probe's MSE as the \textbf{probe loss}.
\end{definition}

\paragraph{Directions, variance, mass, and dead weight.}
The \textbf{variance along a direction} $v$ (a unit vector in activation space) is the variance of $v^\top a$ across the dataset. For $k = 1$, a probe's direction is its normalized weight vector, oriented so the readout increases with the target, and two directions are compared by the angle between them. For $k > 1$, the methods cannot be compared direction by direction, because DAS (\cref{sec:interventions}) finds only a subspace, with no preferred basis within it. We therefore compare the subspaces spanned by each method's directions, by their \textbf{principal angles}\footnote{Principal angles generalize the angle between lines to subspaces: the first is the smallest angle achievable between a unit vector from each, and each subsequent one repeats the construction in the orthogonal complement of the vectors already chosen. Intuitively, if all principal angles are $0^\circ$ the two subspaces coincide, and if all are $90^\circ$ they are orthogonal.} \citep{bjorck1973}, reporting the larger angle. On the HMM testbed of \cref{sec:hmm_setup}, the subspace spanned by the probe's weight vectors is a plane, since the three belief coordinates sum to one. A probe's \textbf{mass} on principal component $i$ of the activations is its share of squared weight there, $m_i = w_i^2/\sum_j w_j^2$, where $w_i$ is the probe's weight along that component.
We define a direction's \textbf{causal operativity} as the mean absolute change in the network's output when the direction's coordinate is interchanged from a source input into a base input, averaged over many such pairs (see \cref{sec:interventions} for specifics). We call a direction \textbf{causally inert} if its causal operativity is near zero, and \textbf{dead weight} for a probe if the probe places large weight on it even though it is causally inert. Whether these directions' covariance with the target is sampling noise or reflects information the network encodes along them is a question we take up in \cref{sec:theory,sec:experiments}.
\begin{definition}[Intervention faithfulness]
\label{def:causal}
Let $g$ be a known ground-truth function specifying the correct network output for any value of the target $f$. Given a probe direction (or, for $k > 1$, the subspace spanned by the probe's weight vectors) and an intervention protocol that shifts the represented value of $f$ by $\delta f$ (\cref{sec:interventions}), the probe is \textbf{intervention-faithful} to the extent that the post-intervention output matches $g(f + \delta f)$: the output shifts as the ground truth predicts. We quantify faithfulness with the \textbf{intervention loss}, $\mathcal{L}_{\text{int}} = \mathbb{E}\big[\|\tilde{y} - g(f + \delta f)\|^2\big]$, where $\tilde{y}$ is the network's post-intervention output and the expectation is over intervention targets and data samples.
\end{definition}

\subsection{Testbed 1: the Mess3 HMM}
\label{sec:hmm_setup}
To test intervention faithfulness, we need a setting where the correct output after an intervention is known exactly. The Mess3 HMM \citep{shai2024} is an ideal testbed for this because the ground-truth latent variable (the belief state) and the correct output for any belief are both exactly computable.
It defines a 3-state hidden Markov model with transition matrix $T_{ij} = P(s_{t+1} = j \mid s_t = i)$ and emission distribution $E_{ik} = P(x_t = k \mid s_t = i)$, parametrized by a mixing parameter $x = 0.05$ and emission concentration $\alpha = 0.85$.
The Bayes-optimal predictor for $P(x_{t+1} \mid x_{1:t})$ must maintain a \emph{belief state} $b_t = P(s_t \mid x_{1:t})$, the posterior over hidden states given observations so far, which is the minimal sufficient statistic for next-token prediction:
\begin{equation}
P(x_{t+1} \mid x_{1:t}) = \sum_{s_t, s_{t+1}} b_t(s_t)\, T_{s_t, s_{t+1}}\, E_{s_{t+1}, x_{t+1}}.
\end{equation}
With 3 hidden states, the belief lives on a 2-simplex. \citet{shai2024} trained a 4-layer transformer (width 64) on this HMM and showed linear probes recover the belief simplex with near-perfect decodability, but did not test the recovered directions causally. We use their pre-trained network, enumerate all sequences of length 1--8, compute each prefix's exact belief and probability, and cache the residual-stream activation at the final LayerNorm on the last token. Probes are fit to the 3-dimensional belief target by weighted least squares, each sequence weighted by its probability under the HMM.
\subsection{Testbed 2: Llama-2-7B on birth year}
\label{sec:llm_setup}
We probe Llama-2-7B for the birth year of named people, using the Wikidata data of \citet{heinzerling2024}. \citet{gurnee2023} showed such numeric attributes are linearly decodable from residual-stream activations, and \citet{heinzerling2024} that they are linearly causally operative. We anchor on this setting precisely because a causally operative direction is known to exist, so a probe's failure to steer implicates the probe, not the absence of a usable direction. Our setup differs in that we fit on the person's true birth year (not the model's expressed value), prompt with the completion ``\texttt{\{name\} was born in 19}'', and read out from the digit logits rather than generated text. The target is shifted and scaled to have mean zero and variance one across the dataset. The probed activation is the 4096-dimensional residual-stream vector at the last token of the name, at layer 8 (selected by a layer sweep of intervention effect, \cref{app:layers}). We use $n = 8000$ prompts. Restricting to 1900--1999 makes the readout a single decade digit. We summarize the output as an \textbf{expected decade}, $\mathbb{E}[d] = \sum_d d \cdot P(d) \in [0, 9]$, from the renormalized next-token distribution over the ten digit tokens.
\subsection{Intervention protocols}
\label{sec:interventions}

\paragraph{HMM: probe-mediated steering.}
Given a probe $W$ and a desired belief shift $\delta f$, we perturb activations by $\Delta a = (W^\top)^{+}\delta f$, where $(W^\top)^{+}$ is the pseudoinverse; these edits span the probe's plane. Interventions move beliefs within the simplex, in 20 smooth steps per target. Intervention loss compares the steered next-token distribution against the belief-implied next-token distribution, averaged over a uniform 496-point grid of target beliefs from three start states, one near each simplex corner.

\paragraph{LLM: natural-magnitude interchange.}
The birth-year target is scalar ($k = 1$), so a probe here is a single direction. Building on \citet{heinzerling2024}, we test a direction by transplanting it from one person to another, so the source's known birth year supplies the ground-truth target. Take a
\emph{base} prompt (say, someone born in the 1920s), a \emph{source} prompt
(someone born in the 1980s), and a candidate unit direction $v$. We input the base
prompt into the network, but interchange the base activation's component along $v$ with the value that
same component takes for the source:
\begin{equation}
\label{eq:interchange}
\Delta a = \left( v^\top a_{\text{src}} - v^\top a_{\text{base}} \right) v .
\end{equation}
If $v$ is intervention-faithful, i.e. the direction the model actually uses for birth year, its predicted decade should move off the
base's and toward the source's. Intervention loss on the LLM is the MSE between the post-interchange expected decade and the source's decade.
\paragraph{Causal operativity.}
To measure a direction's causal operativity, we apply the interchange of \cref{eq:interchange} and record the resulting output change, as the L1 distance between next-token distributions on the HMM and as the absolute change in expected decade on Llama-2-7B.

\paragraph{DAS.}

Distributed Alignment Search \citep{geiger2024das} learns directions by gradient descent on intervention outcomes, and serves as the causal reference point against which probe-derived directions are compared. On Llama-2-7B it learns a single unit direction, randomly initialized, trained to minimize the MSE between the model's differentiable expected decade after the interchange and the source's decade. On the HMM it learns an orthogonal rotation of activation space and interchanges two learned rotated coordinates, matching the two degrees of freedom of the belief simplex, minimizing the MSE against the Bayes-optimal next-token distribution for the source belief.
\section{Theory: Why OLS Probes Find Intervention-Unfaithful Directions}
\label{sec:theory}
To explain why OLS probes find intervention-unfaithful directions, we work in the principal-component (PC) basis of the centered activation matrix and ask what
weight a regression probe places on each principal direction. Let $v_1, \dots, v_d$ be
the principal directions and write $z_j^{(i)} = v_i^\top a_j$ for the coordinate of sample $j$'s
activation along $v_i$, so that each activation is $a_j = \sum_i z_j^{(i)} v_i$. Fix one
principal direction $v$ and drop the superscript: $z_j = v^\top a_j$ is that direction's
coordinate on sample $j$, and $f_j$ is the centered target; for $k > 1$ the analysis applies
to each target coordinate separately, with $\sigma_f$ the standard deviation of the coordinate
in question. Write
$\sigma_v^2 = \frac{1}{n}\sum_j z_j^2$ for the sample variance along $v$ and
$\mathrm{cov}_{vf} = \frac{1}{n}\sum_j z_j f_j$ for the sample covariance.
On Llama-2-7B the $n$ prompts are independent draws from a distribution $D$ over
name-year prompts, and population quantities are expectations under $D$. On the HMM, our dataset contains every sequence of length 1 to 8, each weighted by its probability. Nothing is sampled, so sample and population quantities coincide for this population.

\subsection{The per-direction weight}
\label{sec:theory_a}
In the PC basis, activation coordinates are uncorrelated across the dataset
($\sum_j z_j^{(i)} z_j^{(l)} = 0$ for $i \neq l$), and all variables are centered
(Definition~\ref{def:probe}), so no bias term is needed and the squared error has no cross
terms between weights: the full objective,
\begin{equation}
\label{eq:fullobjective}
\mathcal{L}(w_1, \dots, w_d) = \sum_j \Big( \sum_i w_i z_j^{(i)} - f_j \Big)^2
+ \lambda \sum_i w_i^2,
\end{equation}
decouples into $d$ independent one-dimensional problems, one per direction, and it
suffices to analyze one at a time. Writing $w$ for the weight on the fixed direction
$v$, the one-dimensional objective is
\begin{equation}
\label{eq:objective}
\mathcal{L}_v(w) \equiv \sum_j \left( w z_j - f_j \right)^2 + \lambda w^2 ,
\end{equation}
the sum of squared prediction errors plus the ridge penalty, equal to the
$w$-dependent part of \cref{eq:fullobjective} up to a constant.
\begin{proposition}[Per-direction ridge regression weight]
\label{prop:weight}
The minimizer of \cref{eq:objective} is
\begin{equation}
\label{eq:weight}
w = \frac{\sum_j z_j f_j}{\sum_j z_j^2 + \lambda} = \frac{\mathrm{cov}_{vf}}{\sigma_v^2 + \lambda/n},
\end{equation}
with $\lambda = 0$ recovering the OLS probe.
\end{proposition}
The same formula holds for every principal direction: a direction's weight depends only on the covariance with the target in the numerator and its variance in the denominator.
\subsection{Target-uncorrelated directions and the OLS blow-up}
\label{sec:theory_b}
\begin{definition}[Target-uncorrelated direction]
\label{as:spurious}
A direction $v$ is \emph{target-uncorrelated} if its population covariance with the target is zero.
\end{definition}
Target-uncorrelatedness does not make the sample covariance zero. On sampled data the products $z_j f_j$ are independent zero-mean terms, so the sample covariance, an average of $n$ of them, has typical magnitude equal to its standard error,
\begin{equation}
\label{eq:se}
\mathrm{std}(\mathrm{cov}_{vf}) = \frac{\sigma_v \sigma_f}{\sqrt{n}} .
\end{equation}
\begin{corollary}[OLS blow-up, ridge regression suppression]
\label{cor:blowup}
For a target-uncorrelated direction on sampled data, the typical magnitude of the OLS probe weight is
\begin{equation}
\label{eq:blowup}
\left| w_{\mathrm{OLS}} \right| = \frac{\left| \mathrm{cov}_{vf} \right|}{\sigma_v^2}
\;\sim\; \frac{\sigma_f}{\sigma_v \sqrt{n}}
\;\xrightarrow[\;\sigma_v \to 0\;]{}\; \infty ,
\end{equation}
growing without bound as the direction's variance shrinks. Ridge regression changes only the divisor, which cannot fall below $\lambda/n$: the weight is at most $n \left| \mathrm{cov}_{vf} \right| / \lambda \sim \sigma_v \sigma_f \sqrt{n} / \lambda$, which vanishes as the direction gets quieter, instead of exploding.
\end{corollary}

In a high-dimensional residual stream, most principal components presumably do not causally encode any given target. The argument above shows that any principal component whose covariance with the target is not small relative to its own variance gets a large weight in unregularized regression. On sampled data, we can even quantify the covariance such a component acquires from sampling noise alone (\cref{eq:blowup}). On exhaustive data there is no sampling noise to quantify, but any covariance that a component nevertheless carries for any other reason would be inflated the same way. We call a direction \emph{spurious} if it is intervention-unfaithful yet carries nonzero sample covariance with the target, whatever that covariance's origin. Target-uncorrelated directions on sampled data are the special case where the origin is sampling noise alone, so the covariance's scale is predictable (\cref{eq:se}). Thus, if low-variance spurious directions indeed exist in the residual stream, we expect them to dominate an OLS probe's direction. In-sample decodability cannot reveal this failure. Held-out decodability need not either, since it is not guaranteed to be poor for OLS, and since decodability does not necessarily go hand in hand with intervention faithfulness (which is not always true, \cref{sec:exp_ols}). 

Note that the described amplification is classical, the textbook instability of OLS along low-variance directions \citep{hoerl1970}. But crucially, the classical concern is held-out prediction, not causation. Held-out decodability (in settings where the probe is not already trained on exhaustive data) only penalizes covariance that fails to replicate, and covariance can persist across samples without the network ever using the associated directions (\cref{sec:exp_ols}).

\begin{remark}[Large $\lambda$ does not maximize variance]
\label{rem:massmean}
One might worry that ridge regression simply favors whatever varies most. As $\lambda \to \infty$, \cref{eq:weight} gives $w \propto \mathrm{cov}_{vf}$: ridge regression converges to the direction of covariance with the target, not the direction of maximum variance. A loud direction unrelated to the target gets zero weight.
\end{remark}
\subsection{Scope, and testing for target-uncorrelatedness}
\label{sec:theory_scope}
We stay agnostic about the source of the small covariance that quiet directions carry on the exhaustive HMM data, and measure it directly rather than derive its scale.
On sampled data, target-uncorrelatedness cannot be checked directly but has two testable signatures: sample covariance no larger than the order of its standard error, and no replication on a disjoint sample of prompts. A third expectation for any spurious direction, whether fully target-uncorrelated or not, is that interchanges along such a direction should not move the output faithfully. A faithful direction violates the third by definition, and should violate the first two as well, since a direction the network uses carries genuine covariance with the target. \Cref{sec:experiments} shows that the OLS probes' mass concentrates on directions that satisfy all three signatures.
Importantly, this argument does not promise that a regularized probe finds the intervention-faithful directions, but rather only that an OLS probe should \textit{not} be expected to. Nor does it assume the directions the network uses are high-variance: wherever they lie, the OLS probe's mass is pulled into low-variance noise, and it is this pull, not any claim about where real signal lives, that the theory establishes.

\section{Experiments}
\label{sec:experiments}
We fit and evaluate probes on our full datasets, with no train/test split. The HMM data is an exhaustive enumeration, and we use 8000 samples for Llama-2-7B, twice the 4096 dimensions of the model, so the OLS solution is unique. Ridge probes are fit over a logarithmic grid of $\lambda$, and we call the one with the lowest intervention loss the \textbf{intervention-optimal} ridge probe.

\subsection{OLS probes decode but are causally inert}
\label{sec:exp_ols}

\begin{table}[!ht]
\centering
\caption{The two testbeds tell similar stories. The OLS probes achieve the lowest probe loss (sampled, not exhaustive, on Llama-2-7B), yet their directions carry orders of magnitude less variance, are near-orthogonal to DAS, and interventions along them barely move network outputs. By contrast, intervention-optimal ridge regression probes are similar to the DAS directions in probe loss, intervention loss, variance, and direction. DAS is technically not a probe as it does not optimize for decodability, but its reported ``probe loss'' is nevertheless that of a linear readout fit on its subspace. Also, the intervention loss of 13.7 for the OLS probe on Llama-2-7B is the same as what one gets from not intervening at all, whereas the ridge regression probe (7.1) and DAS (6.4) roughly halve it.}
\label{tab:results}
\footnotesize
\begin{tabular}{llcccc}
\toprule
Testbed & Direction & Probe loss (MSE) & Int.\ loss (MSE) & Variance ($v^\top \Sigma v$) & $\angle(\cdot, \text{DAS})$ \\
\midrule
Mess3 HMM & OLS probe & $9.46 \times 10^{-5}$ & $2.14 \times 10^{-1}$ & $1.16 \times 10^{-8}$ & $89.8^\circ$ \\
 & Ridge probe ($\lambda = 0.1$) & $3.99 \times 10^{-3}$ & $1.95 \times 10^{-3}$ & $5.10 \times 10^{0}$ & $28.9^\circ$ \\
 & DAS & $3.21 \times 10^{-3}$ & $1.88 \times 10^{-3}$ & $9.72 \times 10^{0}$ & $0^\circ$ \\
\midrule
Llama-2-7B & OLS probe & $6.71 \times 10^{-2}$ & $1.37 \times 10^{1}$ & $2.10 \times 10^{-2}$ & $84.7^\circ$ \\
\quad (birth year) & Ridge probe ($\lambda = 10^4$) & $2.51 \times 10^{-1}$ & $7.07 \times 10^{0}$ & $3.83 \times 10^{0}$ & $26.3^\circ$ \\
 & DAS & $6.89 \times 10^{-1}$ & $6.42 \times 10^{0}$ & $8.28 \times 10^{0}$ & $0^\circ$ \\
\bottomrule
\end{tabular}
\end{table}

\begin{figure}[tbp]
\centering
\includegraphics[width=\linewidth]{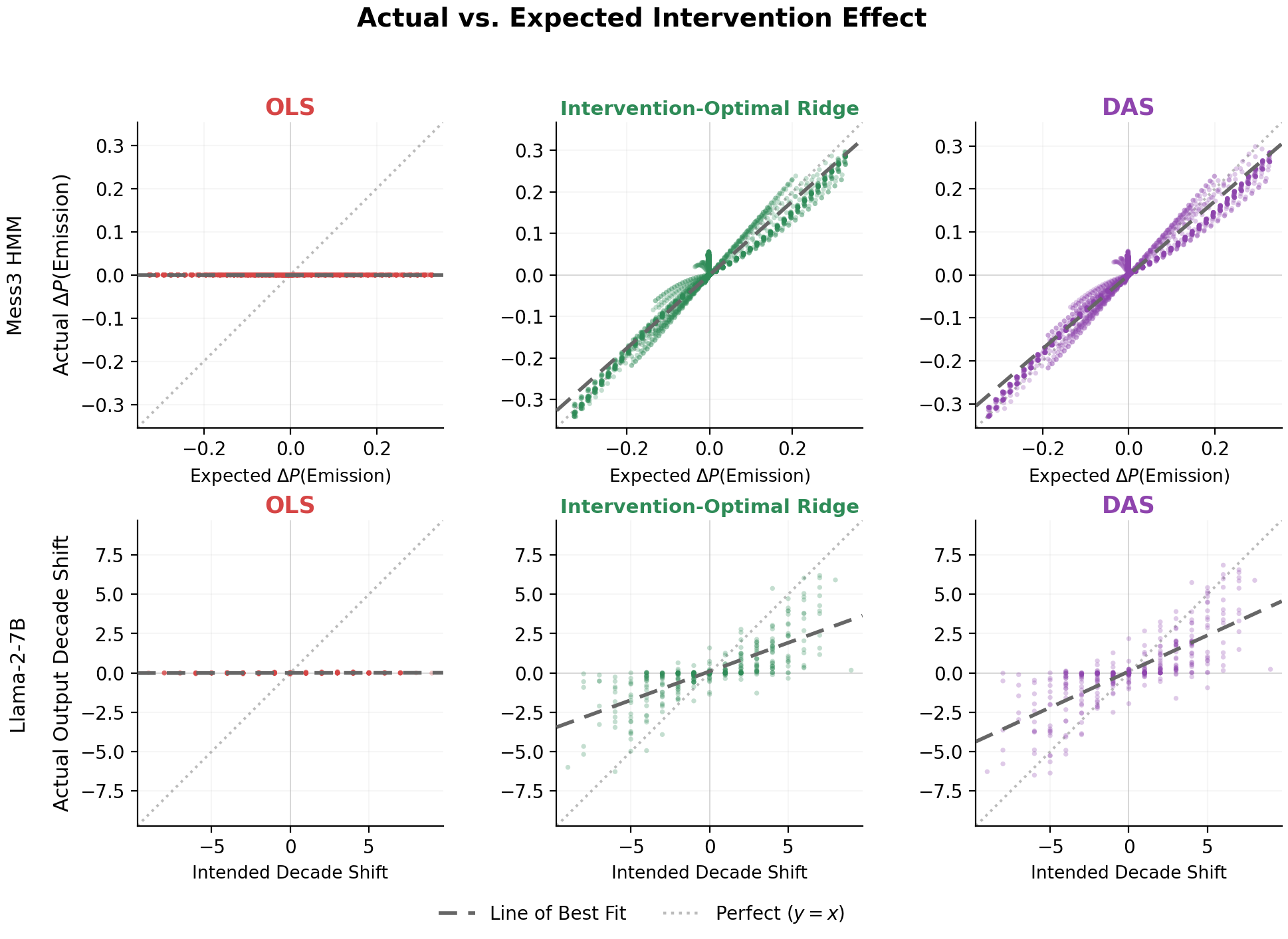}
\caption{Network output change under intervention ($y$) against the change the ground truth predicts ($x$). On the HMM (top) we steer 20 random start beliefs toward each simplex corner at 20 strengths, and each steering step gives three dots, one per emission probability. On Llama-2-7B (bottom) we run 400 base-source interchanges, and each gives one dot, the shift in expected decade. A faithful direction puts dots on the diagonal, an inert one on a flat line at zero. The dashed lines are the best fits. OLS probes are causally inert whereas intervention-optimal ridge regression probes are mostly intervention-faithful, and similar in intervention effect to DAS.}

\label{fig:interventions_grid}
\end{figure}
\begin{figure}[tbp]
\centering
\includegraphics[width=\linewidth]{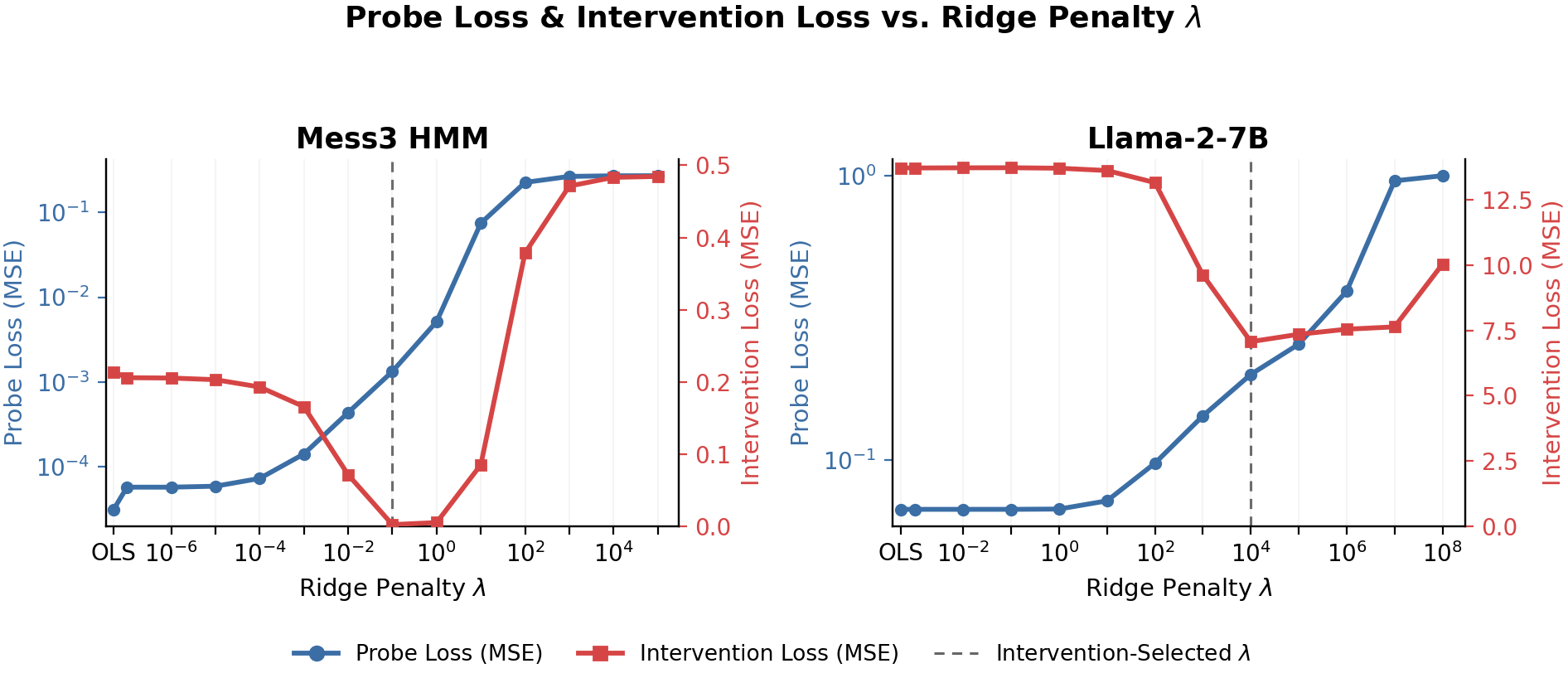}
\caption{Probe loss (blue, log scale) and intervention loss (red) as a function of the ridge penalty $\lambda$. Probe loss rises monotonically with $\lambda$ while intervention loss has an interior minimum.}
\label{fig:sweeps}
\end{figure}
\begin{figure}[tbp]
\centering
\includegraphics[width=\linewidth]{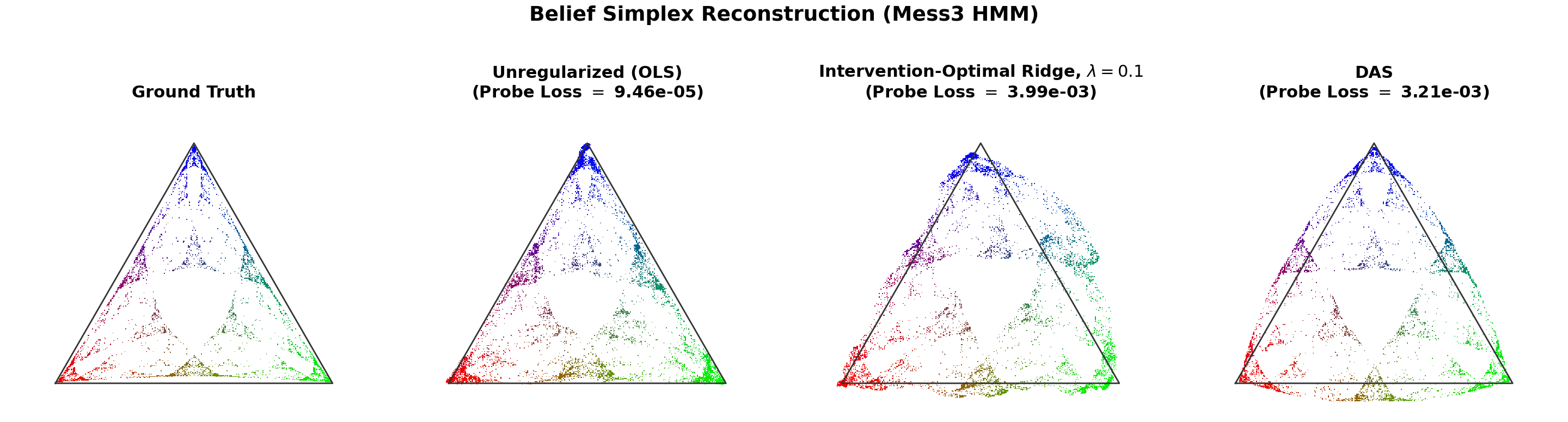}
\caption{Belief simplex reconstructions for ground truth, the OLS probe, intervention-optimal ridge regression probe and DAS. The OLS probe recovers the simplex near-perfectly yet is causally inert. Although the specific distortion shapes differ, the two intervention-faithful methods both produce visibly distorted reconstructions with similar probe loss (more than a magnitude higher than the OLS probe).}
\label{fig:hmm_simplexes}
\end{figure}
\begin{figure}[tbp]
\centering
\includegraphics[width=0.72\linewidth]{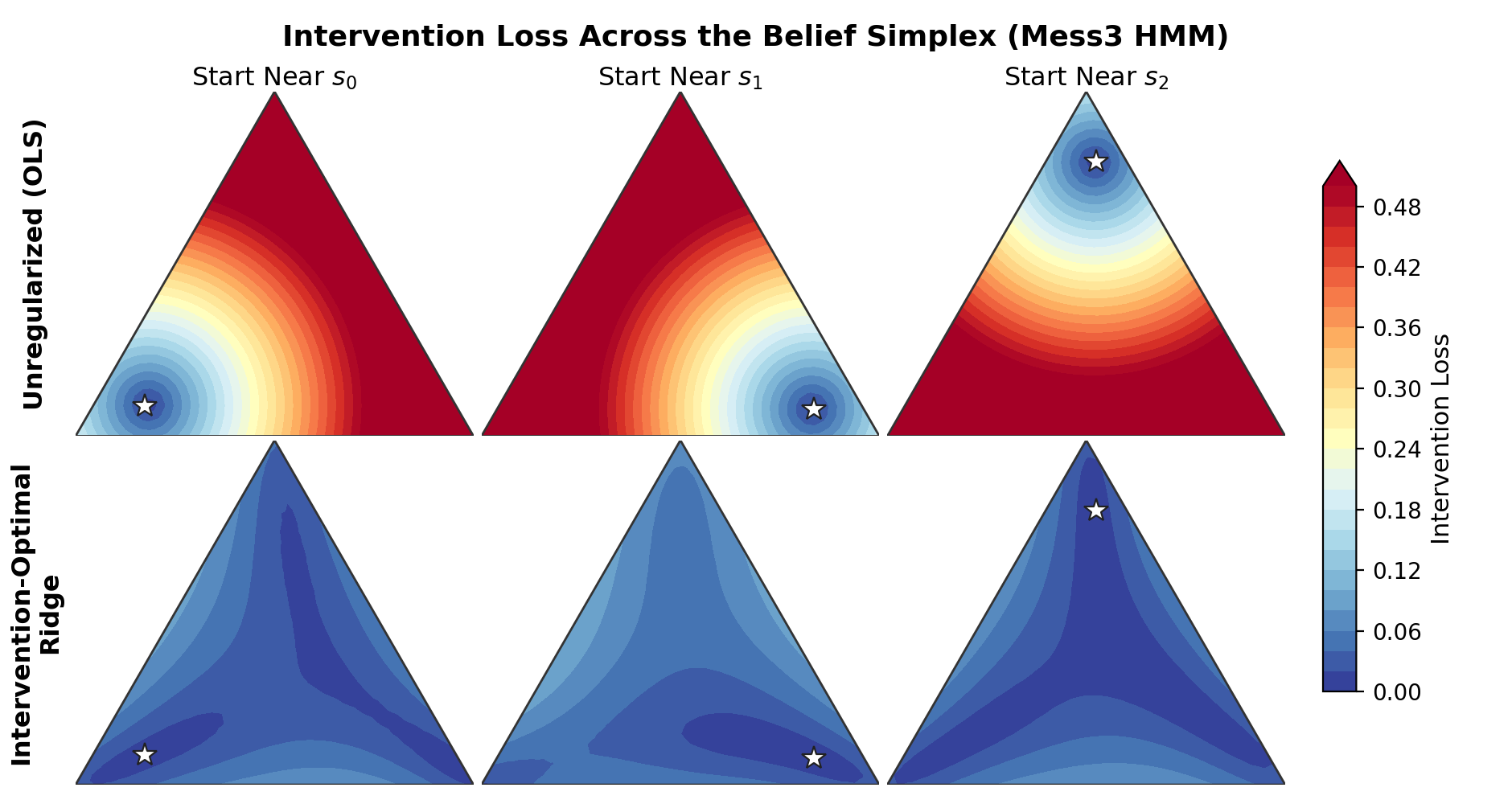}
\caption{Intervention loss across the belief simplex for the OLS probe (top) vs.\ the intervention-optimal ridge regression probe (bottom). From each of the three start states (stars), OLS probe interventions have substantially higher loss than intervention-optimal ridge probe interventions.}
\label{fig:hmm_heatmap}
\end{figure}

On both testbeds, the OLS probe decodes better than any alternative in sample, yet is causally inert, and the variance along its direction is far below the alternatives'.
Held out on an 80/20 split, the Llama OLS probe's loss jumps from $0.05$ on its training split to $0.38$, while intervention-optimal ridge holds ($0.27$ to $0.28$). Held-out decodability thus flags the OLS probe on Llama-2-7B, but not on the HMM. On a held-out HMM split, the OLS probe still decodes an order of magnitude better than the intervention-optimal ridge probe ($4.3 \times 10^{-4}$ against $4.9 \times 10^{-3}$), and is still inert. These held-out numbers are the only ones in the paper. Every other HMM result uses probes fit on the exhaustive enumeration, where there is nothing to generalize to. In that setting, OLS probes trivially achieve higher decodability than intervention-optimal ridge regression probes. Non-trivially, steering with the OLS probe has virtually no effect on the output, while the ridge probe steers mostly faithfully.

\subsection{Intervention-selected ridge regression probes are faithful and high-variance}
\label{sec:exp_ridge}
Intervention-optimal ridge probes are fairly faithful on both testbeds, and their directions carry several orders of magnitude more variance than the OLS directions (\cref{tab:results}).
\subsection{Ridge regression probes rotate onto DAS as \texorpdfstring{$\lambda$}{lambda} grows}
\label{sec:exp_das}
Each ridge probe is fit on decodability alone, with causal data used only to pick $\lambda$, whereas DAS is trained on causation alone.
Yet \cref{tab:results} shows the intervention-selected ridge directions sit close to the intervention-trained DAS directions ($26.3^\circ$ on birth year, $28.9^\circ$ on the HMM), while OLS is nearly orthogonal to them ($84.7^\circ$ and $89.8^\circ$).
\cref{fig:lambda_rotation} shows what happens in between: In the Llama-2-7B experiment, as $\lambda$ grows, the ridge probe direction turns steadily from the OLS direction toward the DAS direction, then degrades again once $\lambda$ gets too large. In the HMM experiment, the intervention-optimal ridge probes are also much closer to the DAS subspace than the OLS probes, but the angle does not grow past the intervention-optimal $\lambda$.
\FloatBarrier
\begin{figure}[t]
\centering
\includegraphics[width=\linewidth]{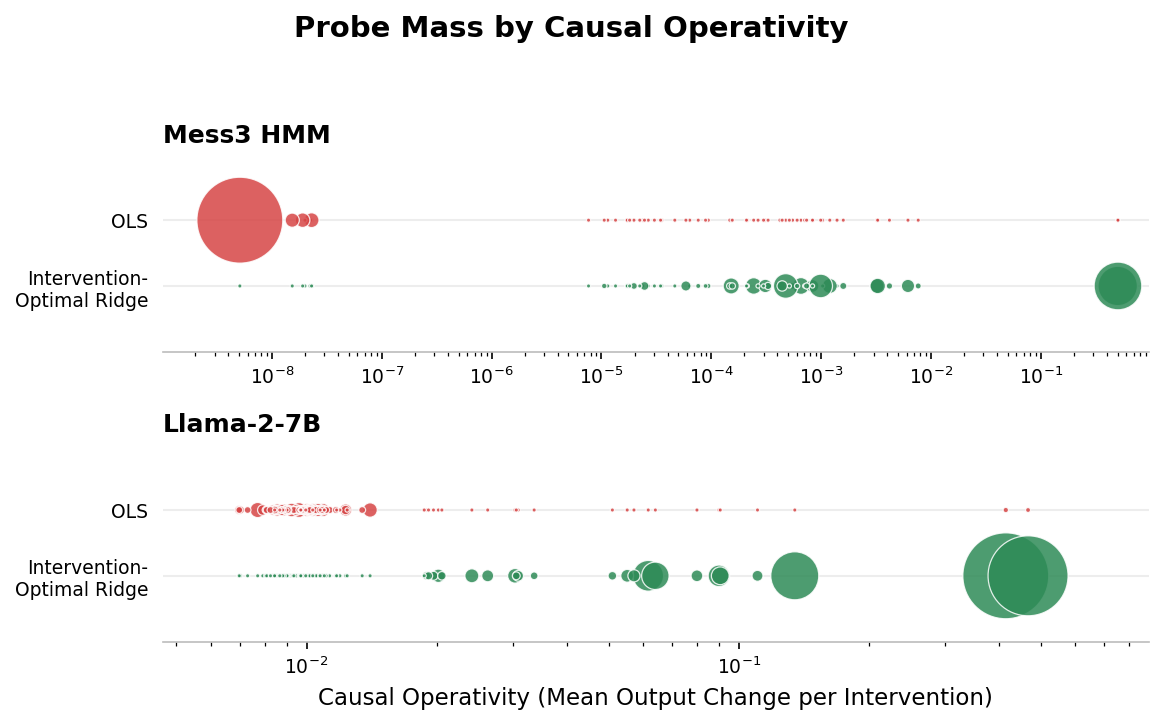}
\caption{Probe mass by causal operativity. Each bubble is a principal component, area proportional to its probe mass $m_i$, position its operativity under in-range swaps. OLS mass (red) concentrates at low operativity on both testbeds; intervention-selected ridge mass (green) concentrates at high operativity. The Llama frame shows the heaviest-mass PCs, covering $95\%$ of the ridge probe's mass but only ${\sim}31\%$ of the OLS probe's, whose remainder spreads over a long tail of small bubbles piling onto the same low-operativity side.}
\label{fig:bubble}
\end{figure}
\subsection{OLS mass sits on dead weight}
\label{sec:deadweight}
A principal component, unlike a probe, has no readout $(W^\top)^{+}$, so ``steer it toward a target value'' is undefined, and the only defined intervention is the interchange itself, whose causal operativity is defined in \cref{sec:interventions}.

\Cref{fig:bubble} shows where each probe's mass sits. On both testbeds the OLS probe's mass lands on directions of near-zero operativity, while the ridge probe's mass lands on strongly operative ones. On the HMM a single direction of near-zero variance ($10^{-18}$) holds $0.915$ of the OLS probe's mass.

\Cref{fig:bubble} tests principal components one at a time, which leaves a loophole. The OLS probe spreads its mass over many components, and a combination of individually inert directions could still be operative. To close it, we let DAS search the span of the $100$ heaviest-weight OLS components on Llama-2-7B (holding $34.8\%$ of the OLS probe's mass) for the most causally effective direction it can find. It finds none that is causally operative, let alone faithful (\cref{app:rdas}).
\section{Discussion}
\label{sec:discussion}
\paragraph{Why the convergence of ridge regression probes and DAS is informative.}
Intervention loss is the objective DAS trained on, so DAS being intervention-faithful is about as surprising as a decodability-trained probe being decodable. The informative part is ridge regression converging onto DAS. Selection cannot create directions, only choose among a dozen pre-computed candidates, each fit on decodability alone. Yet one lands within $26$--$29^\circ$ of DAS, much closer than the OLS direction ($85$--$90^\circ$). 

\paragraph{Ridge regression is not simply chasing variance.}
One might object that the intervention faithfulness of ridge regression probes comes simply from favoring high-variance and that the most intervention-faithful direction (or subspace, on the HMM) should be the highest-variance one. On Llama-2-7B, this is not what we find. The intervention-optimal ridge direction has variance $3.83$ and largely ignores the top principal component, whose variance is $25{,}570$, about $6{,}700$ times larger. Note that variance cannot replace the causal test in selecting $\lambda$ as variance is monotone in $\lambda$, so maximizing it causes over-regularization.

\paragraph{Scope and limits.}
Causal control is partial. Even the most intervention-faithful ridge probe directions leave intervention loss well above zero, though far below that of the OLS probes (\cref{fig:interventions_grid}). Further, the LLM results use only one model. Also, this theory covers regression probes on continuous targets, whereas most probing in practice is logistic. 

\paragraph{OLS probes' intervention unfaithfulness does not prove ridge probes' faithfulness.}
Corollary~\ref{cor:blowup} proves only the negative claim that an OLS probe is prone to spurious covariance between low-variance directions and the target. It does not prove the positive claim that ridge regression finds the directions the network uses, nor that intervention-faithful directions are necessarily high-variance. That the intervention-optimal ridge direction is intervention-faithful and similar to DAS is an empirical finding from the two testbeds (\cref{sec:exp_ridge,sec:exp_das}). The recommendation to regularize rests on those experiments, as well as the theoretical result that ridge regression removes a failure mode OLS is proven to have, even though ridge regression is also not guaranteed to find intervention-faithful directions.

\paragraph{Related work: statistics and logistic probing.}
Ridge regression is classically justified by generalization, shrinking weights to predict better on new data \citep{hoerl1970}. Ours is a claim about causal use, not prediction, and the 80/20 split on the HMM separates the two cleanly. There the OLS probe generalizes very well (held-out probe loss $4.3 \times 10^{-4}$) and is still causally inert. Further, as mentioned before, the concept of held-out decodability does not even apply when fitting on the exhaustive HMM dataset. Rather, ridge regression here \textit{buys intervention faithfulness at the expense of decodability}.   

Anchor regression \citep{rothenhausler2021anchor} also argues that regularization is useful for finding causal structure, although using a wholly different argument than ours. The paper models data linearly (such as hourly bike rentals predicted from temperature, humidity, and wind) and penalizes the part of the fit that changes between days. Their reason, borrowed from \citet{peters2016}, is that weights that change from day to day cannot be the underlying causal ones.   

RAPTOR \citep{raptor2026} also sweeps the $L_2$ penalty for probes (``ridge-adaptive logistic probes''), but selects by validation accuracy, not intervention effect. Intervention-selected sweeps for logistic probing, and a logistic counterpart of the theory, are natural next steps.

\paragraph{Related work: interventions and concurrent work.}
\citet{makelov2024} caution that interventions can mislead when they push activations to values that never occur naturally, activating pathways the network never uses. We limit this risk by construction: interventions request only belief shifts within the simplex, and the LLM interchange copies activation values the source actually exhibits. \citet{modell2026} steers a nonlinear manifold of time representations in Llama-2-7B, showing that a year representation can be steered. The work compares no probing methods causally and exhibits no decodable-but-inert direction, which is our focus.

\section{Conclusion}
\label{sec:conclusion}
We have shown that ordinary least squares regression probes tend to point in low-variance directions of activation space that carry small but nonzero covariance with a feature target. Because this leaves decodability intact not just in-sample, but also when fitting on exhaustive datasets and sometimes even out-of-sample, the resulting directions can appear reliable but fail miserably under causal intervention. Experiments on both a toy model transformer and Llama-2-7B show that OLS probes are causally inert, whereas ridge regression probes with the penalty selected by intervention effect are intervention-faithful. We also find that the latter occupy high-variance directions and that they align closely with the intervention-trained, decodability-blind directions found by Distributed Alignment Search. These results provide a mechanistic justification for regularizing regression probes. %

\section*{Acknowledgments} Our two testbeds build directly on the experimental setups of \citet{shai2024} and \citet{heinzerling2024}, and we thank the authors for making their work publicly available.

\newpage
\appendix
\crefalias{section}{appendix}
\crefalias{subsection}{appendix}
\section{Layer Sweep}
\label{app:layers}
\begin{figure}[H]
\centering
\includegraphics[width=0.85\linewidth]{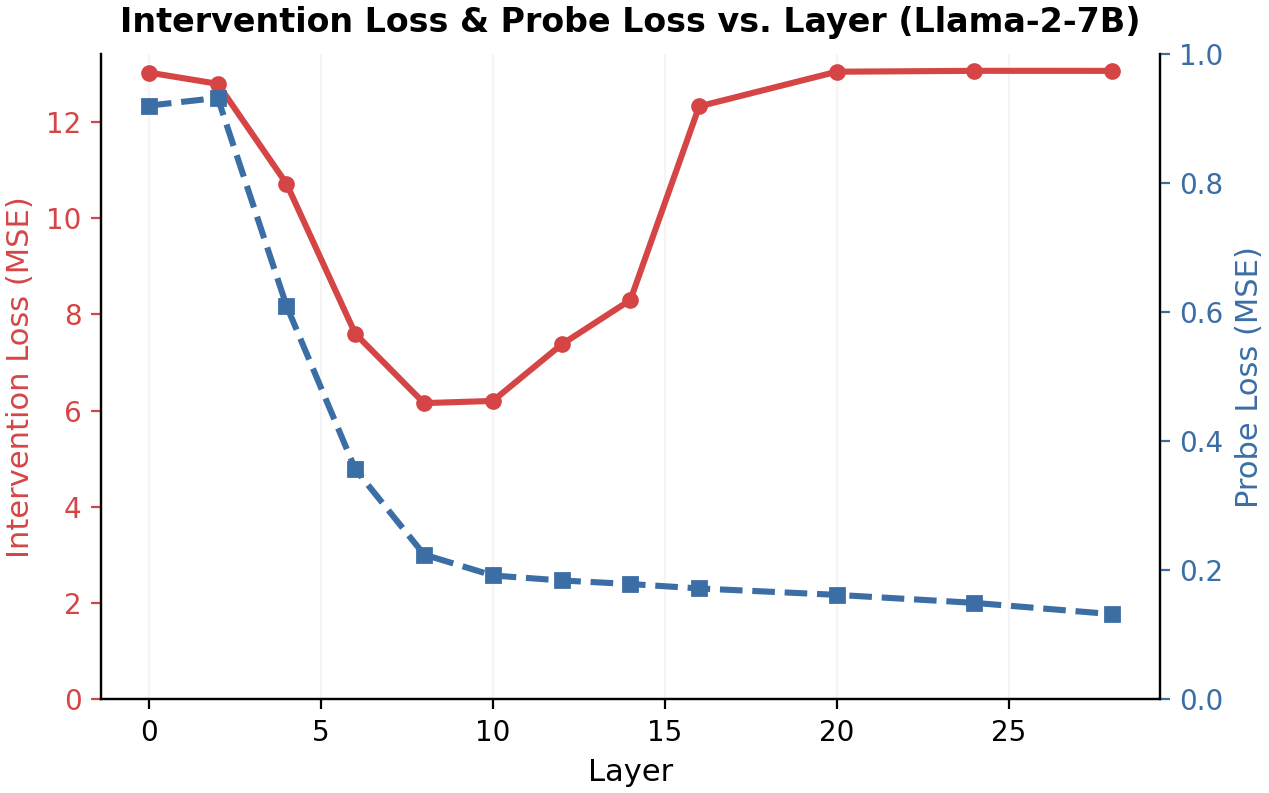}
\caption{Intervention loss (red, MSE) and probe loss (blue, MSE) of the birth-year interchange by layer, based on 200 base–source pairs. Each layer has its own intervention-optimized ridge regression probe. Interventions work best at layer 8, the site used for our Llama-2-7B experiments.}
\label{fig:layer_sweep}
\end{figure}

\section{Ridge-to-DAS Rotation Across the Sweep}
\label{app:rotation}
\begin{figure}[H]
\centering
\includegraphics[width=0.9\linewidth]{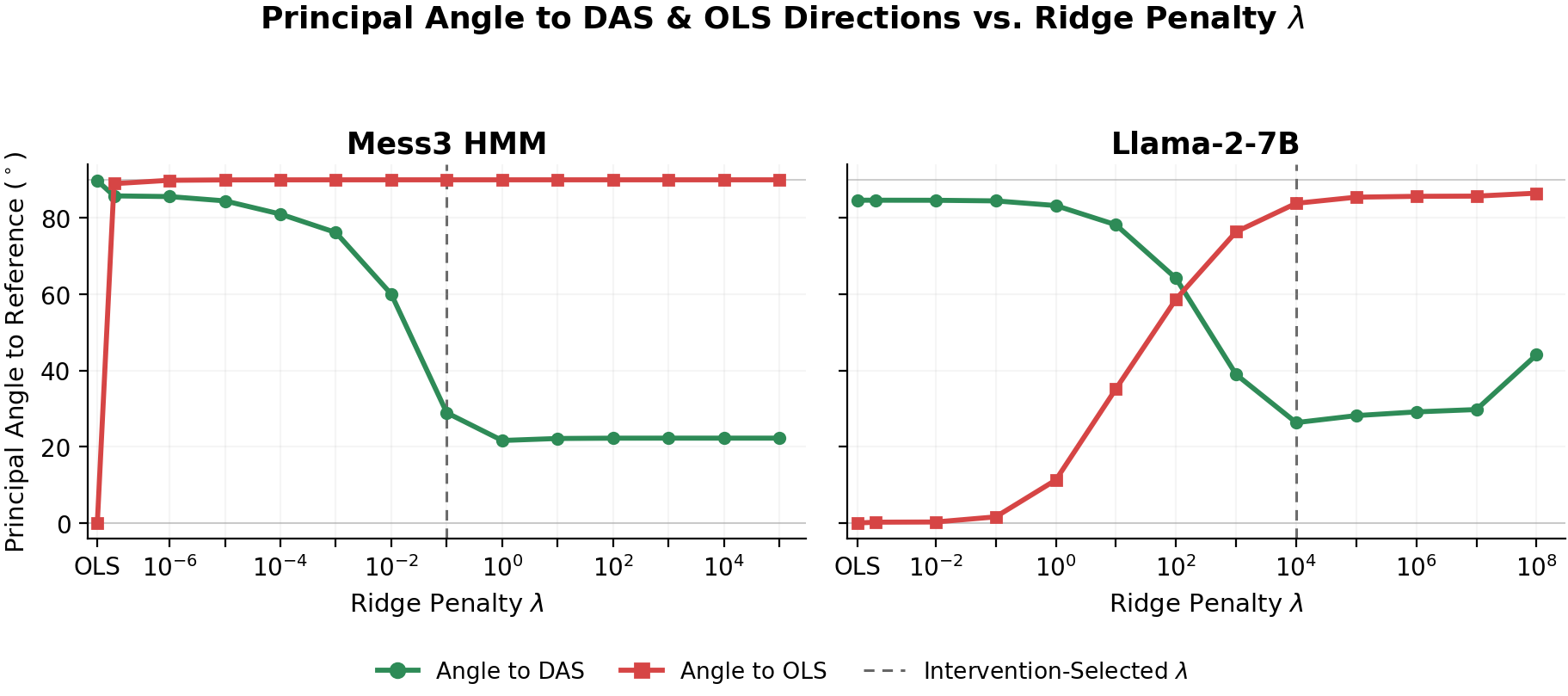}
\caption{Angle from the ridge probe to DAS (green) and to OLS (red) as $\lambda$ grows. On the HMM study, we report the larger of the two principal angles between the steering subspaces. On the Llama-2-7B study, the angle is between single directions. The intervention-selected $\lambda$ is dashed. The sweep's positive case for ridge regression is clearest on the Llama-2-7B experiment, where the principal angle to DAS is minimized at the intervention-optimal $\lambda$. The case against OLS is clear on both testbeds.}
\label{fig:lambda_rotation}
\end{figure}
\section{Covariance at the Predicted Scale}
\label{app:zhist}
\begin{figure}[H]
\centering
\includegraphics[width=.9\linewidth]{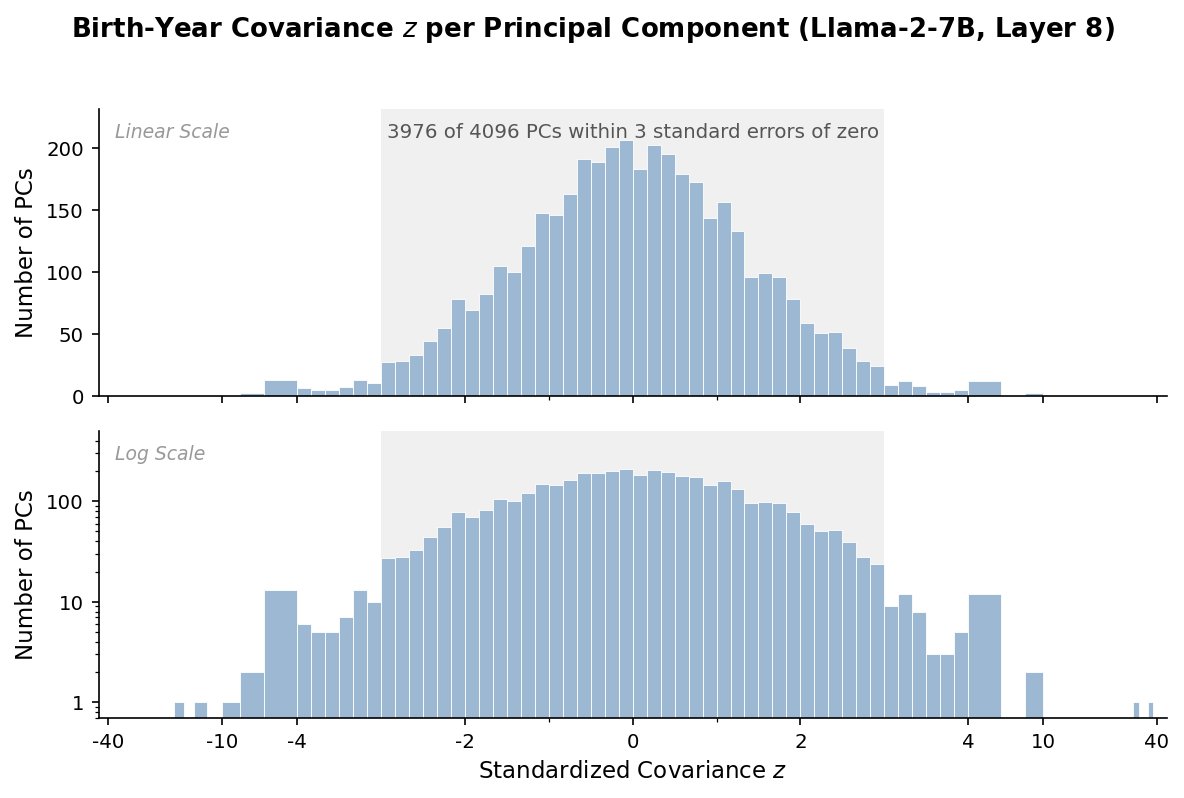}
\caption{Standardized birth-year covariance $z_i = \sqrt{n}\,\mathrm{cov}_i/(\sigma_i \sigma_f)$ for all $4096$ PCs of Llama-2-7B layer 8, computed split-half (one half of the prompts fixes the PC directions, the other supplies $\sigma$ and $\mathrm{cov}$). $3976$ of $4096$ PCs lie within $3$ standard errors of zero (shaded), the sampling-noise scale of \cref{sec:theory_b}.}
\end{figure}
\section{Restricted DAS over the Heaviest OLS Directions}
\label{app:rdas}
\begin{figure}[H]
\centering
\includegraphics[width=.85\linewidth]{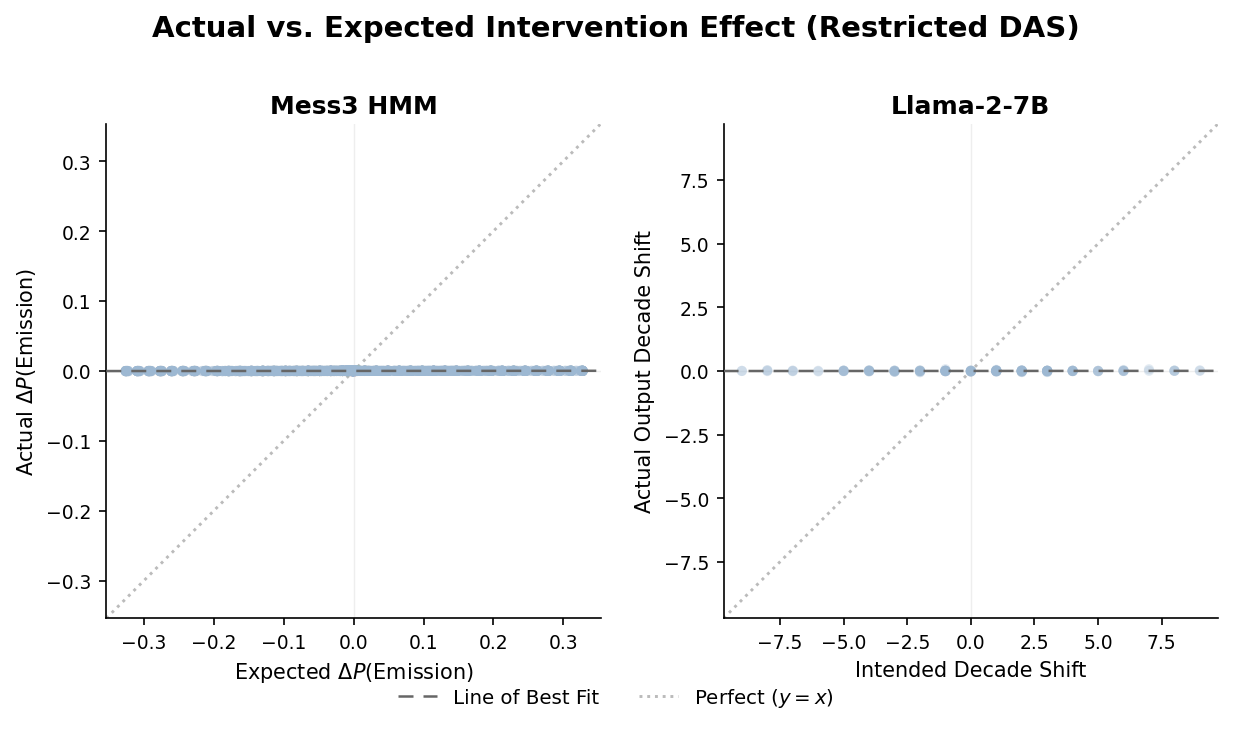}
\caption{Per-sample intervention effect for DAS confined to the heaviest-OLS-weight principal components: $100$ on Llama-2-7B and $7$ on the HMM transformer (over $99.9\%$ of OLS probe mass). No causally operative direction is found in either.}
\end{figure}
\end{document}